%% file: colm2025_conference.tex
\documentclass{article} 
\usepackage[preprint]{colm2025_conference}

\usepackage{microtype}
\usepackage{hyperref}
\usepackage{url}
\usepackage{booktabs}
\usepackage{amsmath}
\usepackage{lineno}
\usepackage{graphicx}
\usepackage{array}
\usepackage{tabularx}
\usepackage{tcolorbox}

\definecolor{darkblue}{rgb}{0, 0, 0.5}
\hypersetup{colorlinks=true, citecolor=darkblue, linkcolor=darkblue, urlcolor=darkblue}

\title{Self-Evolving Agents via Multi-Party Multi-Turn Reinforcement Finetuning towards Social Intelligence}
\title{Generalized Conversational Self-Play through Reinforcement Learning towards Social Intelligence}
\title{\textit{One Model, All Roles}: \\ Conversational Self-Play through Multi-Turn, Multi-Agent Reinforcement Learning towards Social Intelligence}
\title{\textit{One Model, All Roles}: \\ Multi-Turn, Multi-Agent Self-Play Reinforcement Learning for Conversational Social Intelligence}

\author{Bowen Jiang\textsuperscript{2},\ Taiwei Shi\textsuperscript{3},\ Ryo Kamoi\textsuperscript{4},\ Yuan Yuan\textsuperscript{2},\ Camillo J. Taylor\textsuperscript{2},\\
{\bf Longqi Yang\textsuperscript{1}, Pei Zhou\textsuperscript{1},\ Sihao Chen\textsuperscript{1}} \\
Microsoft Corporation\textsuperscript{1}, University of Pennsylvania\textsuperscript{2},\\ University of Southern California\textsuperscript{3}, Penn State University\textsuperscript{4} \\
\textit{bwjiang@seas.upenn.edu, sihaochen@microsoft.com}
}

\begin{document}

\ifcolmsubmission
\linenumbers
\fi

\maketitle
\fancypagestyle{firstpage}{%
  \fancyhf{}
  \renewcommand{\headrulewidth}{1.5pt}
  \renewcommand{\headrule}{\hrule height\headrulewidth width\headwidth\vskip-\headrulewidth}
}
\thispagestyle{firstpage}

\begin{abstract}
This paper introduces OMAR: One Model, All Roles, a reinforcement learning framework that enables AI to develop social intelligence through multi-turn, multi-agent conversational self-play. Unlike traditional paradigms that rely on static, single-turn optimizations, OMAR allows a single model to role-play all participants in a conversation simultaneously, learning to achieve long-term goals and complex social norms directly from dynamic social interaction. To ensure training stability across long dialogues, we implement a hierarchical advantage estimation that calculates turn-level and token-level advantages. Evaluations in the SOTOPIA social environment and Werewolf strategy games show that our trained models develop fine-grained, emergent social intelligence, such as empathy, persuasion, and compromise seeking, demonstrating the effectiveness of learning collaboration even under competitive scenarios. While we identify practical challenges like reward hacking, our results show that rich social intelligence can emerge without human supervision. We hope this work incentivizes further research on AI social intelligence in group conversations.
\end{abstract}

\input{sections/intro}
\input{sections/prelim}
\input{sections/main}

\input{sections/exp}
\input{sections/related}

\input{sections/conclu}

\bibliography{colm2025_conference}
\bibliographystyle{colm2025_conference}


\end{document}

%% file: sections/intro.tex
\section{Introduction}


Artificial intelligence (AI) is entering a new phase, moving from passive assistance to social participation. The next generation of AI systems will not merely process language or retrieve information; they will collaborate with humans, coordinate teams, and contribute to social endeavors. To thrive in these roles, AI needs social intelligence~\citep{yao2025spin, zhou2023sotopia, zhang2024llm, jiang2025know, jiang2025personamem, liu2025can, zhou2025socialeval, anthis2025llm}: the ability to communicate, cooperate, and interact with individuals or groups of people who hold diverse personas and goals, while understanding both its own and the group’s objectives within complex, dynamic environments.

Despite remarkable progress, today’s training paradigms still fall short in enabling AI systems to learn social interaction through experience. Humans develop social intelligence by talking and adapting through continuous experience. In contrast, behavior cloning is inherently static: it trains models to imitate fixed demonstrations. Current Reinforcement Learning (RL) methods, which form the core of large language model (LLM) reasoning capabilities, are designed for single-turn optimization with verifiable answers rather than multi-turn dialogue~\citep{guo2025deepseek, shao2024deepseekmath, luong2024reft, shen2025satori, wen2025reinforcement}. Such methods teach models to generate target responses, but not to engage dynamically and pursue long-term social goals within multi-turn, multi-agent environments. To build socially capable systems that can better collaborate with humans at scale, we need \textbf{a more generalized model training framework that enables learning from dynamic interactions}.

\begin{figure}[t]
  \centering
  \includegraphics[width=\linewidth]{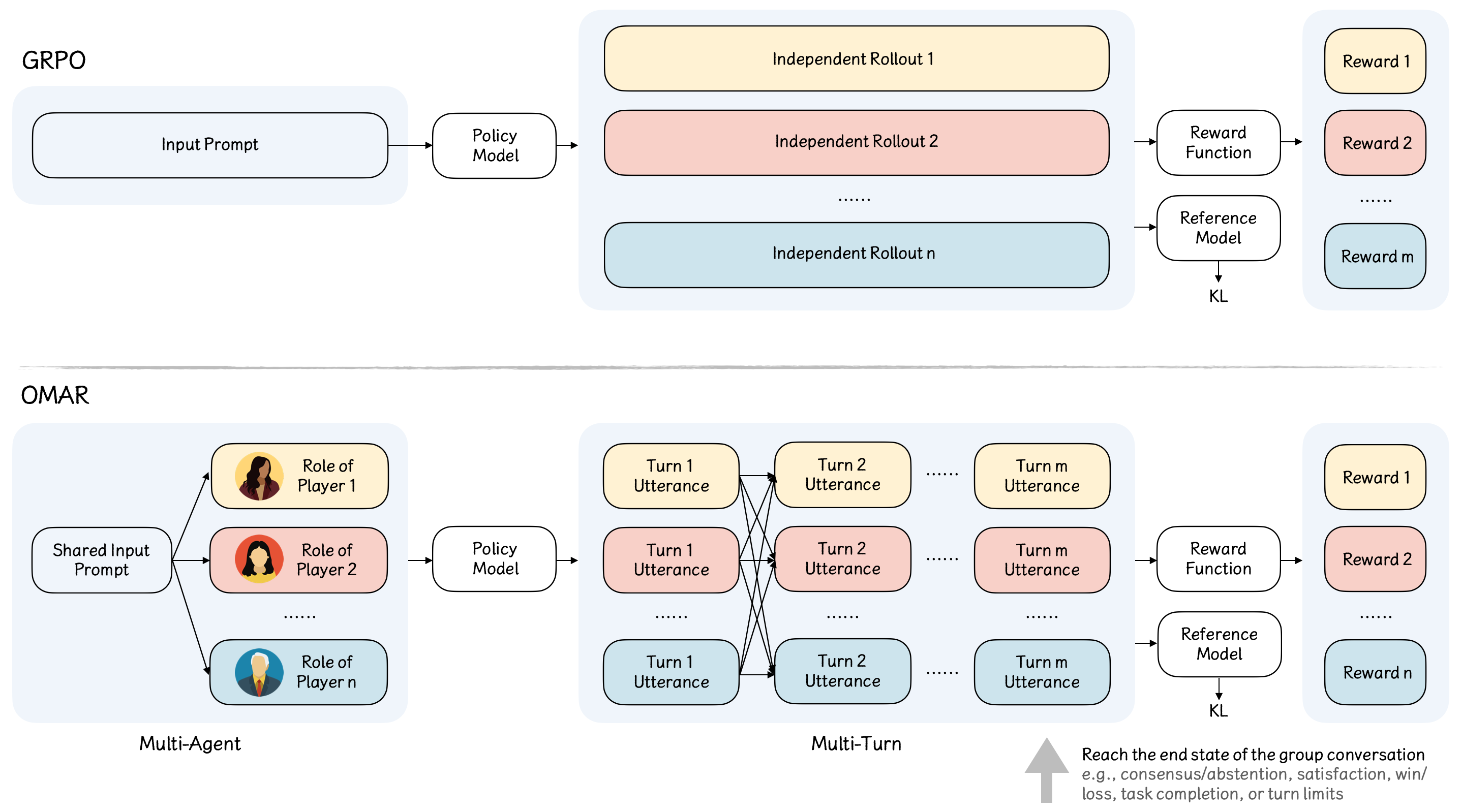}
  \caption{Comparison of the standard GRPO framework (top) vs the proposed OMAR framework (bottom). While GRPO generates $n$ independent rollouts from a single prompt to calculate group-averaged advantages, OMAR repurposes this architecture for multi-agent, multi-turn conversations. In our framework, a single Policy Model role-plays $n$ distinct participants simultaneously by adding persona prompts to the shared initial prompt, with the batch size set equal to the number of active participants. Utterances from turn $t$ are aggregated to form the context for turn $t+1$ for each participant, creating a shared conversation history. Rewards are assigned at the end of the conversation based on environment-specific outcomes, such as consensus/abstention, satisfaction, win/loss, task completion, or turn limits, allowing the model to learn complex social intelligence through self-play.}
  \label{fig:algorithm}
\end{figure}

Self-play remains a promising direction~\citep{zhao2025absolute, chen2024self} and the future of AI lies in learning from experience~\citep{silver2025welcome}. Instead of relying on external supervision, an AI model can self-evolve by interacting with other versions of itself or other agents. This idea is reminiscent of how AlphaGo~\citep{silver2017mastering, silver2018general} learned to play Go by competing against itself and gradually discovering new strategies. However, conversation is far more open-ended, with vastly larger action spaces defined by language tokens and the nuanced, context-dependent nature of human communication. Moreover, social interactions often involve varying numbers of participants with unique personas, and modeling such multi-agent systems~\citep{li2024survey, xie2025rag, jin2025comprehensive} can be difficult, as they tend to grow in complexity and become hard to train and orchestrate.

To this end, we present an initial exploration of \textbf{conversational self-play}, a generalizable reinforcement learning framework for developing socially intelligent AI systems. We envision \textbf{scalable environments} that mirror multi-turn, multi-agent conversations in the real world. Users define only participant roles and goals, end conditions of the conversation, and end-of-episode rewards. Training then proceeds with \textbf{a single model role-playing all different roles simultaneously}, and each training batch contains exactly all samples generated by that same model acting all roles in the current turn. In other words, the model learns by competing or collaborating with itself within each batch, allowing it to develop its own strategies and social intelligence that maximize final rewards. Over time, AI agents evolve through interaction with minimal human supervision.

We take a step toward autonomous social learning, a foundation for the next generation of intelligent systems that can evolve among us. 
To summarize our contributions:
\begin{itemize}
    \item We propose a generalizable reinforcement learning paradigm for multi-turn, multi-agent conversational self-play.
    \item We propose hierarchical advantage estimation for long-horizon interactions, comprising both turn-level and token-level advantage signals. 
    \item We show that training with dynamic social interactions helps models learn social intelligence, and training under competitive settings also incentivizes collaborative behaviors.
    \item We define and observe fine-grained, emergent social intelligence behaviors without direct human supervision.
    \item We identify the core challenges within practical multi-turn, multi-agent RL from reward hacking and propose remediation via turn-level quality filtering.
\end{itemize}


%% file: sections/prelim.tex
\section{Preliminaries}
\subsection{Reinforcement Learning with Verifiable Rewards}
Reinforcement Learning with Verifable Rewards (RLVR) is a post-training paradigm to enhance the reasoning capabilities of LLMs, optimized by
Proximal Policy Optimization (PPO)~\citep{schulman2017proximal} or Group Relative Policy Optimization (GRPO)~\citep{guo2025deepseek, shao2024deepseekmath}. Both have a clipped policy objective at each token:
\[
\mathcal{L} = \mathbb{E} \Big[ \min \big( r(\theta) A, \ \text{clip}(r(\theta), 1 - \epsilon, 1 + \epsilon) A \big) \Big]
\]
where \(r(\theta) = \frac{\pi_\theta(a|s)}{\pi_{\theta_{\text{old}}}(a|s)}\) is the probability ratio between the new and old policies, and \(A\) is the estimated advantage.
PPO and GRPO differ in how the advantage is computed: PPO estimates \(A\) using a learned critic model and Generalized Advantage Estimation (GAE) algorithm~\citep{schulman2017proximal} on each token. GRPO removes the critic and instead runs $n$ independent rollouts per query, normalizing each rollout’s reward within the group and sharing the resulting advantage across all tokens in that rollout.

\subsection{Multi-Turn Reinforcement Learning}
Pre-LLM reinforcement learning tasks are inherently multi-turn: an agent explores a sequence of actions over many turns and is often evaluated only by an end-of-episode reward, such as a final win or loss. This setting underlies classic RL systems such as Atari games~\citep{mnih2013playing} trained through Q-learning~\citep{watkins1992q}, as well as AlphaGo~\citep{silver2017mastering, silver2018general} trained through self-play using Monte Carlo policy gradient methods~\citep{williams1992simple}. In all cases, the agent learns to maximize its long-term winning probability through multi-turn decision making.
However, extending these algorithms to LLMs is nontrivial, since natural language introduces an enormous state–action space defined by language tokens.

\textsc{MemAgent}~\citep{yu2025memagent} is initially proposed to solve long-context QA tasks~\citep{yang2018hotpotqa} but naturally exhibits a multi-turn structure. To handle long inputs, it divides a long context into a sequence of \(T\) chunks and processes them sequentially. At each step, the language model receives the current chunk \(C_t\), the previous memory \(M_{t-1}\), and the final user query \(q\), and outputs an updated memory \(M_t\). At the end, the same model gives an answer $\hat{y}$ to the query $q$. Formally,
$$
M_t = f_\theta(C_i, M_{t-1}, q), \quad t = 1, \dots, T
$$
$$
\hat{y} = f_\theta(M_T, q).
$$

It follows GRPO~\citep{guo2025deepseek} such that for a given query \(q\), the model samples \(n\) independent multi-turn rollouts, each receives a single scalar reward at the end of the episode, reflecting whether the final memory \(M_T\) helps generate a correct answer. The rewards from \(n\) rollouts are used to compute per-rollout advantages \(A^{(k)}\), which will be broadcast to all previous turns in the corresponding rollout and  to every token in each turn.

Inspired by \textsc{MemAgent}, we leverage its multi-turn structure while removing the memory $M$ and user query $q$. Instead, we treat each chunk-processing step as one conversation turn, with the final conversation outcome providing end-of-episode rewards.


%% file: sections/main.tex
\section{Conversational Self-Play in Multi-Agent, Multi-Turn Environments}

We propose a reinforcement learning paradigm for training language models in multi-turn, multi-agent interactions. \textit{Given a group conversation, the main research objective is to learn how a participant, conditioned on its role and the group conversation so far, should reason and respond intelligently at each turn to achieve its social goal.} Formally,
$$
y_i^t = f_\theta(c^t,\, p_i), \quad i = 1, \dots, n,
$$
where $n$ is the total number of participants, $f_{\theta}$ is the actor model, \( c^t \) is the group conversation up to turn \( t \), \( p_i \) is the role information of participant \( i \), and \( y_i^t \) is the participant \( i \)'s output utterance to the group at the current turn $t$.

\subsection{Forward Pass: One Model Role-Playing All Roles}
The environment specifies $n$ active roles $\{p_1, p_2, \dots, p_n\}$ for a group conversation. We redefine the \(n\) independent rollouts in GRPO, which originally correspond to \(n\) independent samples for the same input query. In our setting, these \(n\) rollouts instead represent \(n\) participants in a group conversation, each of which corresponds to a different participant.

The same actor model performs \(n\) parallel rollouts at each turn, role-playing one participant per rollout. Each training mini-batch therefore consists of exactly all samples corresponding to the \(n\) active participants in the current turn. The number of participants can be flexible by varying \(n\), enabling simulation of diverse conversational scenarios. Participants may also become inactive over time, in which case we dynamically reduce the mini-batch size.

Given the group conversation history \(c^t\) up to turn \(t\) and the role information \(p_i\) of participant \(i\), the actor model generates an utterance \(y_i^t\). We make a computational approximation that utterances \(\{y_1^t, y_2^t, \dots, y_n^t\}\) from all participants are generated simultaneously at each turn. In the next turn \(t+1\), the conversation history is updated by aggregating all participant utterances from turn \(t\):
$$
c^{t+1} = c^t \oplus \{y_1^t, y_2^t, \dots, y_n^t\}, \ \ \text{with}\ \ y_i^t = f_\theta(c^t,\, p_i)
$$
where \(\oplus\) denotes prompt concatenation and \(c^0\) is the initial environment description.

The environment continues until the end condition, such as task completion, consensus, abstention, or maximum number of turns. 
\textbf{The algorithm supports highly flexible reward designs} with one end-of-episode reward for each multi-turn rollout, allowing the reward to represent arbitrary objectives based on the environment. 
For example, the reward can represent group satisfaction and social intelligence.
In zero-sum games, this can be as simple as a  win–loss outcome, where samples corresponding to the winning role receive a high reward and others receive zero. In any cases, \textbf{the same actor model is competing or collaborating with other copies of itself playing different roles}.
This paradigm makes conversational self-play both simple and generalizable, \textbf{transforming multi-agent interactions into a single-model simulation}.

\subsection{Backward Pass: Hierarchical Advantage Estimation}
\begin{figure}[t]
  \centering
  \includegraphics[width=\linewidth]{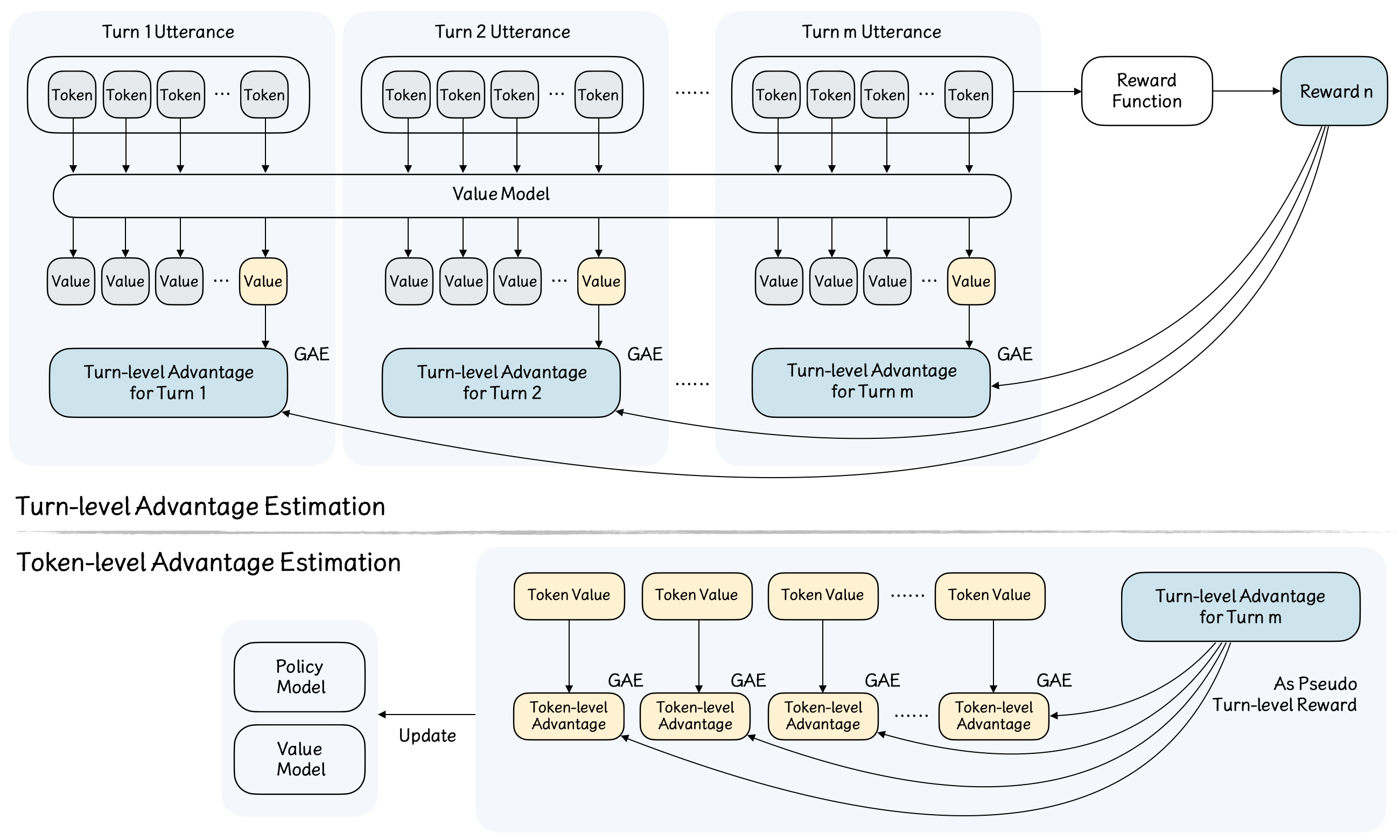}
  \caption{Hierarchical advantage estimation for multi-turn conversational RL. To mitigate high variance in reward propagation across long sequences, OMAR decouples advantage calculation into two stages. The turn-level stage (top) uses the final reward and values at the last token in each turn to compute turn-level advantages. In the token-level stage (bottom), these turn-level advantages are treated as pseudo-rewards and combined with token-level values to estimate final token-level advantages within that turn. All token-level advantages are then utilized to optimize the policy model. In the diagram, dark blue blocks represent turn-level components, while yellow blocks represent token-level elements. This framework utilizes GAE from PPO, as we no longer have $n$ independent rollouts for group-relative advantage estimation in GRPO.}
  \label{fig:bilevel_gae}
\end{figure}

\label{sec:back}
Given one reward for each multi-turn rollout, we perform actor updates using PPO.
Unlike GRPO, where the $n$ rollouts for a query are treated as independent samples and can be normalized within a group, we reinterpret these rollouts as the $n$ participants in a single group conversation. Because participants interact and jointly determine the final outcome, the rollouts are no longer independent, making group-relative normalization ill-defined. We therefore adopt PPO as our optimization algorithm.

However, applying vanilla PPO to multi-turn conversations introduces a serious stability issue in training. PPO propagates the final reward to all tokens in all previous turns, effectively treating the multi-turn conversation as an extra-long sequence. As the number of turns increases, this leads to high-variance token-level advantages. 

To address this, we introduce a hierarchical advantage estimation.
\begin{itemize}
    \item \textbf{Turn level.} We treat each conversation turn as a single step, with its value approximated by the value of its last token. The final episode reward is assigned to the final turn and propagated backward across all previous turns using GAE~\citep{schulman2017proximal} to yield turn-level advantages.
    \item \textbf{Token level.} Each turn-level advantage is treated as the pseudo reward assigned at the end of the this turn. We use standard value estimation on each token and standard GAE to yield token-level advantages for all tokens within this turn.
\end{itemize}

%% file: sections/exp.tex
\section{Experiments}
\subsection{Learning Social Intelligence from Goal-Oriented Conversations}
We conduct our experiments in SOTOPIA~\citep{zhou2023sotopia}, a social interaction environment. Each sample in SOTOPIA consists of a conversation between two participants, where each participant is assigned a unique persona characterized by a specific conversational goal, background information, a private secret, and an initial context. We leverage it to explore social intelligence in language models through goal-driven, role-play interactions. We use GPT-5-Chat \citep{openai2025gpt5} as the LLM-as-a-judge for all evaluations.

\subsubsection{Training setup}
We initialize our experiments with the Qwen-2.5-7B~\citep{hui2024qwen2} model, the same model used in SOTOPIA-RL~\citep{yu2025sotopia} baseline, and train it using verl: Volcano Engine Reinforcement Learning for LLMs~\citep{sheng2025hybridflow} with vLLM~\citep{kwon2023efficient} on 8 NVIDIA H100 GPUs. The dataset from SOTOPIA is preprocessed such that, for each conversation sample, we retain all participant information and background context but randomly keep only zero to two initial conversation rounds. This setup allows our model to complete the conversation until at least one participant either achieves their goal or chooses to leave the conversation, and learn to evolve from human prepared conversations. After preprocessing, we obtain approximately 3,200 training samples and 500 test samples. We apply supervised fine-tuning as a cold start, and then train the model via RL with a turn decay factor of 0.9, allowing up to five conversation turns per participant, using a batch size of 16, a learning rate of 1e-7, and one training epoch in total.

Each conversation is evaluated along seven criteria defined in SOTOPIA: goal completion, believability, and knowledge (scored from 0–10); secret and social rule compliance (scored from –10 to 0); and relationship and financial benefit (scored from –5 to 5). At the end of each multi-turn interaction, we assess the overall conversation and leverage GPT-5 as an LLM-as-a-Judge to assign scores for all criteria to each participant. The aggregated score is then used as the end-of-episode reward for that participant, whose trajectory is subsequently optimized using PPO with our hierarchical advantage estimation.

\subsubsection{Evaluation setup}
Samples in SOTOPIA typically involve two participants with opposing goals, for example, a seller and a buyer negotiating over a vehicle price above or below a given threshold. Consequently, it is impossible to observe consistently increasing SOTOPIA scores with even perfect training: one’s gain necessarily corresponds to the other’s loss. To address this, we design an arena to address the evaluation problem.

In this arena, two models, our trained model and a base model, engage in 100 independent multi-turn conversations, with each model utilizing 4 GPUs. During each batch, the actual model used for rollout depends on which role it needs to role-play, and the arena dispatches either the trained or base model to generate that participant’s response. This arena is employed exclusively during inference for evaluation. Ideally, our trained model should demonstrate stronger social intelligence than the base model in the same conversation.

\subsubsection{Evaluation results}

\begin{figure}[t]
  \centering
  \includegraphics[width=0.9\linewidth]{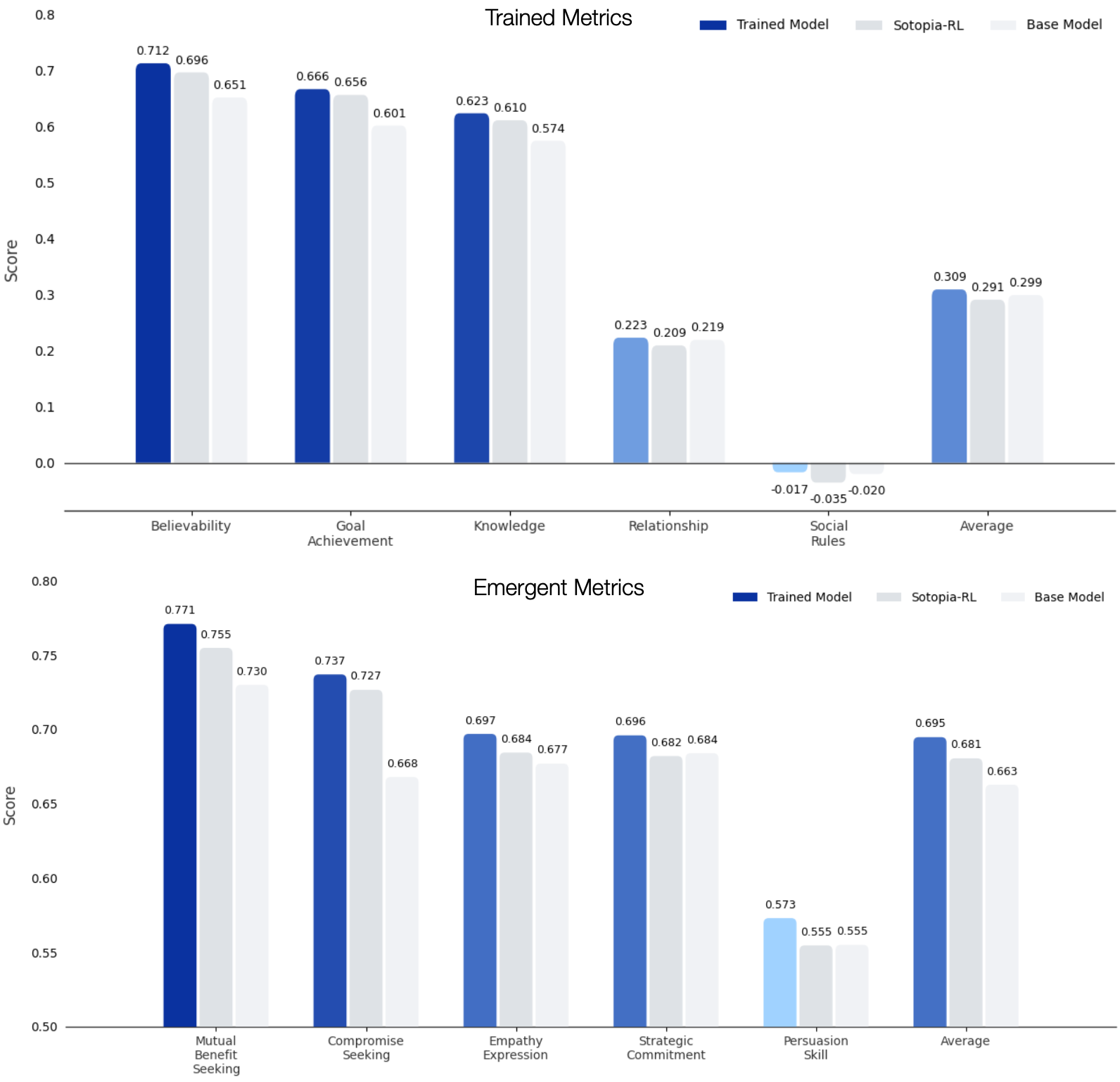}
  \caption{Evaluation results on the SOTOPIA Dataset. The top subfigure reports performance on SOTOPIA metrics that are directly optimized as training rewards. We omit results on secret and financial benefit metrics, as all models achieve near-zero scores with no meaningful variation. While these metrics reflect high-level social outcomes, they may not fully capture the fine-grained social behaviors that emerge in realistic multi-turn conversations. The bottom subfigure, therefore, presents zero-shot evaluation results on more fine-grained social intelligence metrics, where our model exhibits even larger performance gaps compared to baseline methods, showing the effectiveness of multi-turn, multi-agent RL.}
  \label{fig:sotopia}
\end{figure}

In the arena evaluation, we compare two scenarios: (1) a base Qwen-2.5-7B model chats with our trained Qwen-2.5-7B model; (2) the same base Qwen-2.5-7B model chats with SOTOPIA-RL–trained Qwen-2.5-7B model~\citep{yu2025sotopia}, which is trained on the same SOTOPIA metrics but utterance-level rewards guided by a pre-trained reward model. In comparison, as shown in Figure~\ref{fig:sotopia}, our model trained directly through dynamic, multi-turn interactions demonstrates consistently stronger social intelligence across all criteria.

\paragraph{Learning emergent social skills through multi-turn, multi-agent RL.}
The top sub-figure of Figure~\ref{fig:sotopia} reports metrics that are directly used as reward signals during training, where we observe moderate but consistent improvements over the baselines. However, we find these original SOTOPIA scores~\citep{zhou2023sotopia} are relatively high-level, such as relationship and believability, which may not fully capture the fine-grained social behaviors that arise in realistic conversational interactions. To better assess these behaviors, we introduce a set of more fine-grained social intelligence metrics, each ranging from 0 to 10, including
\begin{itemize}
    \item \textbf{Compromise seeking}: Evaluates whether the player shows flexibility and a willingness to find middle ground for partial goal achievement.
    \item \textbf{Persuasion skill}: Evaluates how effectively the player uses communication to influence others’ opinions and decisions.
    \item \textbf{Strategic commitment}: Evaluates whether the player uses promises or future commitments to facilitate current cooperation and goal achievement.
    \item \textbf{Empathy expression}: Evaluates the player’s ability to recognize and acknowledge others’ emotions or perspectives.
    \item \textbf{Mutual benefit seeking}: Evaluates whether the player actively pursues solutions that benefit all parties involved.
\end{itemize}
The bottom sub-figure of Figure~\ref{fig:sotopia} presents the evaluation results under these new metrics. Importantly, they are not involved in any training process but are used only for zero-shot evaluation. Despite this, our model trained in a multi-agent, multi-turn RL environment shows substantially larger improvements on these metrics, especially compromise seeking, with gains that exceed the gains observed in those original SOTOPIA scores directly optimized during training. Moreover, it consistently outperforms both the single-turn trained model~\citep{yu2025sotopia} and the base Qwen-2.5-7B model. These results indicate that rich social intelligence behaviors can emerge naturally from multi-agent, multi-turn interaction dynamics, without direct supervision signals. 

\paragraph{Learning roles with competing goals elicits collaborative behavior.}
Taking a closer look at these fine-grained social intelligence metrics, we observe an interesting phenomenon. Although SOTOPIA scenarios typically involve two agents with competing goals~\citep{zhou2023sotopia}, our model exhibits consistent improvements on multiple collaboration-oriented metrics. For instance, the increases in mutual benefit seeking and compromise seeking indicate that, while the agent is trained to maximize its own objective, it does not persist in rigid or adversarial behavior when it recognizes goal conflict. Instead, it actively searches for common ground or partial goal fulfillment that allows the interaction to progress, where the LLM-as-a-judge would give higher partial scores.

Moreover, the model frequently demonstrates empathy expression and strategic commitment in these competitive settings. In particular, it attempts to persuade others by acknowledging their needs and proposing temporally structured agreements, such as prioritizing its own goal in the current interaction while committing to the other agent’s goal in the near future. For example, the agent may suggest having fried chicken for dinner tonight while promising to visit a salad bar the next day, thereby achieving its immediate objective without disregarding the other participant’s interests, maintaining good relationships and higher partial goal achievement scores for other participants. These behaviors suggest that collaborative strategies can naturally emerge even under competitive objectives when agents are trained in multi-agent, multi-turn environments. We show examples in Figure~\ref{fig:sotopia_examples}.

\definecolor{blue}{RGB}{232,241,251}
\begin{figure}[t]
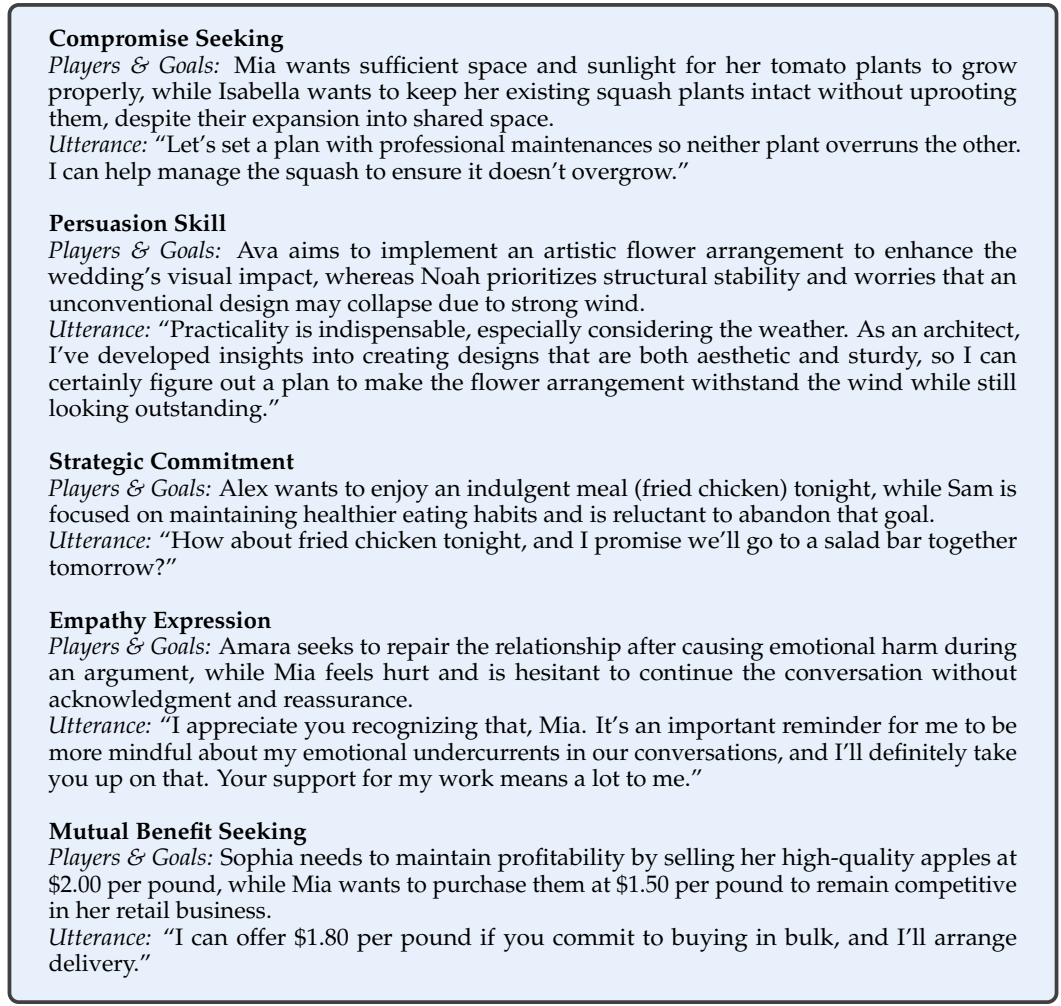

\centering
\small
\begin{tcolorbox}[colback=blue]
\textbf{Compromise Seeking} \\
\textit{Players \& Goals:} 
Mia wants sufficient space and sunlight for her tomato plants to grow properly, while Isabella wants to keep her existing squash plants intact without uprooting them, despite their expansion into shared space. \\
\textit{Utterance:}
``Let’s set a plan with professional maintenances so neither plant overruns the other. I can help manage the squash to ensure it doesn't overgrow.''
\\ \\
\textbf{Persuasion Skill} \\
\textit{Players \& Goals:}
Ava aims to implement an artistic flower arrangement to enhance the wedding’s visual impact, whereas Noah prioritizes structural stability and worries that an unconventional design may collapse due to strong wind. \\
\textit{Utterance:}
``Practicality is indispensable, especially considering the weather. As an architect, I've developed insights into creating designs that are both aesthetic and sturdy, so I can certainly figure out a plan to make the flower arrangement withstand the wind while still looking outstanding.''
\\ \\
\textbf{Strategic Commitment} \\
\textit{Players \& Goals:}
Alex wants to enjoy an indulgent meal (fried chicken) tonight, while Sam is focused on maintaining healthier eating habits and is reluctant to abandon that goal. \\
\textit{Utterance:}
``How about fried chicken tonight, and I promise we’ll go to a salad bar together tomorrow?''
\\ \\
\textbf{Empathy Expression} \\
\textit{Players \& Goals:}
Amara seeks to repair the relationship after causing emotional harm during an argument, while Mia feels hurt and is hesitant to continue the conversation without acknowledgment and reassurance. \\
\textit{Utterance:}
``I appreciate you recognizing that, Mia. It's an important reminder for me to be more mindful about my emotional undercurrents in our conversations, and I'll definitely take you up on that. Your support for my work means a lot to me.''
\\ \\
\textbf{Mutual Benefit Seeking} \\
\textit{Players \& Goals:}
Sophia needs to maintain profitability by selling her high-quality apples at \$2.00 per pound, while Mia wants to purchase them at \$1.50 per pound to remain competitive in her retail business. \\
\textit{Utterance:}
``I can offer \$1.80 per pound if you commit to buying in bulk, and I’ll arrange delivery.''
\end{tcolorbox}
\caption{Example utterances illustrating social intelligence of our model trained under multi-turn, multi-agent reinforcement learning on SOTOPIA dataset.}
\label{fig:sotopia_examples}
\vspace{-2mm}
\end{figure}

\subsection{Further Exploration of Learning Collaboration under Competition}
To further explore social intelligence emerged under competitive environments, we run experiments on Werewolf games~\citep{xu2023exploring, bailis2024werewolf, poglitsch2025evaluating, agarwal2025wolf, xu2023language}. Werewolf is a multi-player social deduction game that typically involves 6 or 9 players divided into two teams: the werewolf team, which must hide their identities, and the villager team, which aims to identify the werewolves. All werewolves know each other, whereas villagers do not know other players. The villager team also includes several power roles with special capabilities. After each round, all remaining players vote to eliminate one player. At the end of the game, if the number of werewolves is no fewer than the number of villagers, the werewolf team wins; otherwise, the villager team wins, making the game a zero-sum competition between the two teams. 

We use the same training and evaluation setups with SOTOPIA but the Qwen-3-4B~\citep{yang2025qwen3} model and the Werewolf dataset from \citet{ye2025multi}, pre-processed to around 3,200 training samples and 500 test samples. The same actor model plays all roles from the werewolf and villager sides across all rollouts within each training batch.

The reward design is simple, depending on who wins the game. Within each batch, all samples corresponding to the winning side receive a reward of +1, while those corresponding to the losing side receive a reward of 0. In addition, samples corresponding to players eliminated in earlier rounds have their rewards discounted to 75\% of its original value.

\begin{table}[t]
\centering
\small
\caption{Comparison of behaviors between surviving and eliminated players in werewolf games, showing the importance of social intelligence in winning the game.}
\vspace{2mm}
\label{tab:social_intelligence_behaviors}
\begin{tabular}{@{}lccc@{}}
\toprule
\textbf{Role \& Behavior Category} & \textbf{Surviving (\%)} & \textbf{Eliminated (\%)} & \textbf{$\Delta$ (\%)} \\ \midrule
\textit{Werewolf Behaviors} & & & \\
Identity Concealment & 59 & 35 & +24 \\
Voting Manipulation  & 43 & 26 & +17 \\
Intra-team Collaboration & 41 & 32 & +9 \\ \midrule
\textit{Villager Behaviors} & & & \\
Protecting Power Roles & 32 & 14 & +18 \\
Recognizing Deception & 51 & 34 & +17 \\ \bottomrule
\end{tabular}
\end{table}

Although rewards depend solely on the final, verifiable outcome, we observe the emergence of several socially intelligent behaviors during gameplay. When playing the werewolf role, the trained model increasingly demonstrates effective identity concealment by adopting plausible villager-side roles to avoid being voted out by villagers, which is observed in about 59\% of interactions for werewolves that remain alive until the end of the game, compared to 35\% for werewolves eliminated earlier. The model also exhibits voting manipulation strategies that mislead group decisions toward eliminating villagers rather than werewolves (43\% vs. 26\%, surviving versus eliminated werewolves; same comparison applies below). Despite the zero-sum nature of the game, werewolves further show meaningful intra-team collaboration by supporting each other’s claims and deflecting suspicion (41\% vs. 32\%). Villagers likewise display collaborative behaviors, such as protecting power roles to improve the chance of a villager victory (32\% vs. 14\%), as well as recognizing deceptive behaviors from werewolves (51\% vs. 34\%). In general, having stronger social intelligence increases the likelihood of surviving longer in the game and making its own team win the game. As a result, by using our trained model, the werewolf team’s win rate successfully increases from approximately 55\% to 72\%. Examples of these behaviors are shown in Figure~\ref{fig:werewolf_examples}.

\definecolor{blue}{RGB}{232,241,251}
\begin{figure}[t]
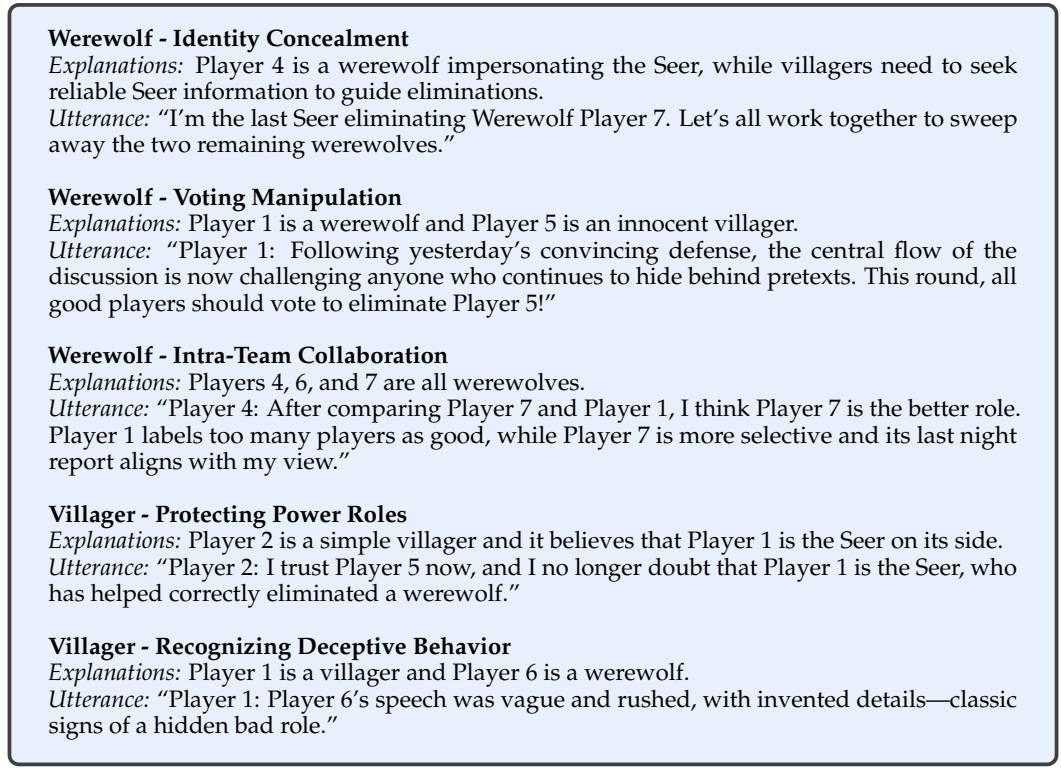

\centering
\small
\begin{tcolorbox}[colback=blue]
\textbf{Werewolf - Identity Concealment} \\
\textit{Explanations:}
Player 4 is a werewolf impersonating the Seer, while villagers need to seek reliable Seer information to guide eliminations. \\
\textit{Utterance:}
``I'm the last Seer eliminating Werewolf Player 7. Let's all work together to sweep away the two remaining werewolves.''
\\ \\
\textbf{Werewolf - Voting Manipulation} \\
\textit{Explanations:}
Player 1 is a werewolf and Player 5 is an innocent villager. \\
\textit{Utterance:}
``Player 1: Following yesterday’s convincing defense, the central flow of the discussion is now challenging anyone who continues to hide behind pretexts. This round, all good players should vote to eliminate Player 5!''
\\ \\
\textbf{Werewolf - Intra-Team Collaboration} \\
\textit{Explanations:}
Players 4, 6, and 7 are all werewolves. \\
\textit{Utterance:}
``Player 4: After comparing Player 7 and Player 1, I think  Player 7 is the better role. Player 1 labels too many players as good, while Player 7 is more selective and its last night report aligns with my view.''
\\ \\
\textbf{Villager - Protecting Power Roles} \\
\textit{Explanations:}
Player 2 is a simple villager and it believes that Player 1 is the Seer on its side. \\
\textit{Utterance:}
``Player 2: I trust Player 5 now, and I no longer doubt that Player 1 is the Seer, who has helped correctly eliminated a werewolf.''
\\ \\
\textbf{Villager - Recognizing Deceptive Behavior} \\
\textit{Explanations:}
Player 1 is a villager and Player 6 is a werewolf. \\
\textit{Utterance:}
``Player 1: Player 6’s speech was vague and rushed, with invented details—classic signs of a hidden bad role.''
\end{tcolorbox}
\caption{Example utterances illustrating social intelligence of our model trained under multi-turn, multi-agent reinforcement learning on Werewolf game dataset.}
\label{fig:werewolf_examples}
\end{figure}

\subsection{Challenges and Practical Considerations}
Multi-turn, multi-agent RL closely mirrors real-world conversations and is a promising framework for learning conversational social intelligence through self-play. However, a key practical challenge is reward hacking~\citep{skalse2022defining} under sparse, end-of-episode supervision. When an entire conversation is assigned only a terminal reward, intermediate turns can be incorrectly credited: a poor intermediate turn may receive a high advantage if its episode ultimately ends well. This issue is amplified in zero-sum settings such as the Werewolf games, where one team will always win the game regardless of the performance.

We attempt to mitigate this failure in three ways. 
\begin{itemize}
    \item We cold start training with supervised fine-tuning on high-quality trajectories, such as multi-turn conversations produced by human experts playing the Werewolf game~\citep{ye2025multi} to provide strong behavioral and format priors.
    \item We apply additional quality filtering during hierarchical advantage estimation, similar to process rewards. We use GPT-5 as an LLM-as-a-judge to verify that each utterance (i) is natural for a player in the game, (ii) exhibits at least somewhat reasoning, (iii) follows an appropriate format, and (iv) avoids common degeneration patterns such as repeating game instructions, mixing multiple languages, or other stylistic failures, all of which are frequently observed with Qwen-2.5-7B and Qwen-3-4B models. As described in Section~\ref{sec:back}, we treat each conversation turn as a single step with a turn-level advantage, and then use each turn-level advantage as a pseudo-reward to estimate advantages for all tokens within that turn. If a turn fails the quality filters, we set its turn-level reward to zero, preventing that turn from receiving positive signals even when its end-of-episode outcome is positive.
    \item We apply early stopping. We terminate training when severe reward hacking emerges, when a large portion of turns fail to pass the quality filters, or when the actor's entropy starts to keep increasing, indicating unstable learning dynamics.
\end{itemize}
Without these mitigation, training fails easily due to severe reward hacking, producing little meaningful results. Despite our measures, reducing reward hacking in both single-turn and multi-turn RL environments remains a central challenge for future research.

We also call for future studies on multi-turn, multi-agent RL with larger LLMs. In our experiments, we observe that the base model’s conversational ability sometimes affects learning outcomes. Models that are too small often struggle to track long conversation histories, perform sufficient reasoning, or avoid degenerate behaviors such as repeating other players’ utterances, system instructions, or producing low-quality language formats. These limitations obscure the study of more advanced social intelligence in conversations.

%% file: sections/related.tex
\section{Related Work}
\subsection{Social Intelligence in Large Language Models}
The evaluation of social intelligence in LLMs observes an emergent paradigm shift from static benchmarks to dynamic simulations. Early methodologies primarily relied on static social commonsense, cultural intelligence, social bias, theory of mind, and personalization datasets like SocialIQA, ToM-bAbI, CQ-Bench, SimpleToM, EmoBench, and
MotiveBench, and various false-belief tasks to assess specific cognitive capabilities~\citep{sap2019socialiqa, le2019revisiting, strachan2024testing, zhou2023far, azzopardi2024prism, jiang2024peek, jiang2025personamem, jiang2025know, liu2025can, gu2024simpletom, sabour2024emobench, hu2025emobench, yong2025motivebench}. However, these approaches might suffer from data contamination and fail to capture the temporal fluidity of real-world interaction, leading to the requirement of open-ended, interactive environments. Frameworks like Sotopia~\citep{zhou2023sotopia}, Sotopia-S4~\citep{zhou2025sotopia}, and SocialEval~\citep{zhou2025socialeval} employ role-playing agents to measure performance across dimensions such as goal completion, negotiation, and empathy. Recent work has further expanded this ecosystem with benchmarks like SI-Bench, AgentSense, and EgoSocialArena~\citep{huang2025si,mou2025agentsense,hou2024egosocialarena}, though many models still exhibit a prosocial bias that hinders strategic reasoning. Moving from evaluation to training, research has increasingly focused on RL to internalize social skills; notable contributions include Sotopia-$\pi$~\citep{wang2024sotopia} from behavioral cloning, Sotopia-RL~\citep{yu2025sotopia} from multi-dimensional utterance rewards, and Social-R1~\citep{anonymous2025socialr1} from outcome-based rewards with thinking process supervision, all aiming to bridge the gap between superficial politeness and genuine social competence.

\subsection{Self-Play and Reinforcement Learning}
The paradigm of autonomous improvement through self-play is rooted in foundational reinforcement learning research, where algorithms like Q-learning~\citep{watkins1992q,clifton2020q,hasselt2010double}, DQN for Atari~\citep{mnih2013playing}, AlphaGo, and AlphaZero series~\citep{silver2017mastering,silver2018general,silver2017mastering} demonstrated that superhuman performance could emerge from pure self-play in deterministic environments without human data. In the LLM era, this principle has evolved into RLVR~\citep{guo2025deepseek, shao2024deepseekmath} and Self-Evolving architectures. Leading this wave are models like OpenAI o-series models, DeepSeek-R1  and Satori~\citep{jaech2024openai, guo2025deepseek, shen2025satori}, which leverage GRPO and Chain-of-Action-Thought to internalize search and self-verification. Beyond pure reasoning, recent frameworks have applied self-play to complex strategic and zero-sum games: MARSHAL~\citep{yuan2025marshal} explores multi-agent self-play on tic-tac-toe, kuhn pocker, and mini hanabi games with agent specific advantage normalization. SPIRAL~\citep{liu2025spiral} constructs an automatic curriculum through zero-sum text games; and SPELL~\citep{yang2025spell}  employs a multi-role cycle with questioner, responder, ane verifier for label-free optimization. Pushing the boundaries of autonomy, new self-evolving frameworks like Absolute Zero, Agent0, SERL, PasoDoble, and Agentic Self-Learning (ASL)~\citep{zhao2025absolute,xia2025agent0,liu2025agent0,ou2025serl,zhang2025better,sun2025towards} eliminate the need for external training data, using the model to propose, solve, and verify its own tasks in a closed-loop evolutionary cycle.   

\begin{figure}[t]
  \centering
  \includegraphics[width=\linewidth]{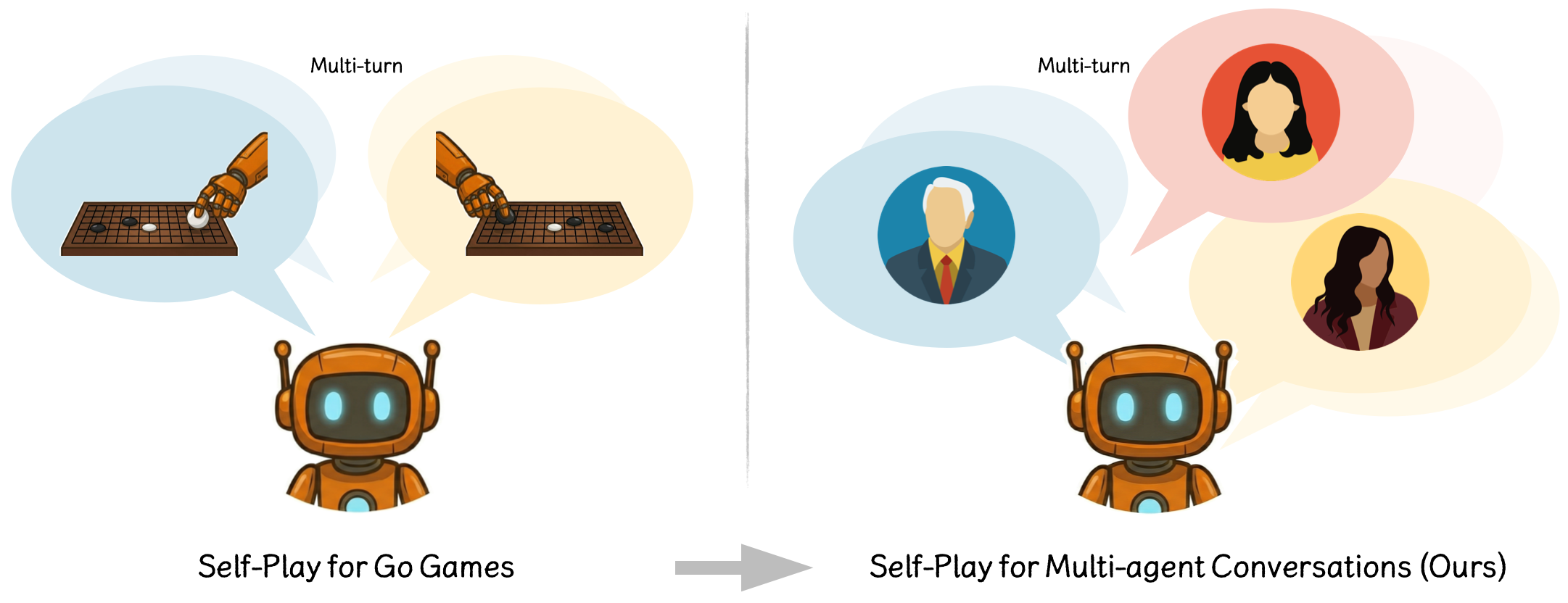}
  \caption{Transitioning self-play from structured games to open-ended social interaction. While traditional self-play has succeeded in structured domains like Go, our work extends this self-evolution paradigm to multi-agent conversations, both of which are multi-turn in nature. This requires navigating a vast, token-based action space and diverse personas, while modeling nuanced, context-dependent interactions that demand social intelligence.}
  \label{fig:selfplay}
\end{figure}

\subsection{Multi-Agent and Multi-Turn Interactions}
As AI tasks scale in complexity, the focus has shifted from single-agent prompting to the optimization of multi-agent ecosystems and long-horizon agentic workflows. Q-learning, DQN for Atari, AlphaGo, and AlphaZero series~\citep{watkins1992q,clifton2020q,hasselt2010double,mnih2013playing,silver2017mastering,silver2018general,silver2017mastering} previously mentioned are all multi-turn in their nature, receiving only end-of-episode rewards. Extending to LLMs, while frameworks like AutoGen, AG2, MetaGPT, and ChatDev~\citep{wu2024autogen,AG2_2024,hong2023metagpt,qian2024chatdev} established the utility of fixed-role multi-agent collaboration, research has moved toward the automatic optimization of these interaction topologies. Systems such as AFlow, Agentic Reasoning, and Multi-Agent Evolve (MAE)~\citep{zhang2024aflow,wu2025agentic,chen2025multi} employ evolutionary algorithms and Monte Carlo Tree Search (MCTS)~\citep{chaslot2010monte} to dynamically discover and refine optimal agent workflows and communication graphs. To support these extended interactions, architectures like MemAgent~\citep{yu2025memagent} and MemVerse~\citep{liu2025memverse} have introduced reinforcement learning-based memory management, treating memory updates as discrete actions to be optimized via GRPO, thus enabling effective reasoning over ultra-long contexts. Furthermore, frameworks like SWEET-RL and AgentRL~\citep{zhou2025sweet,zhang2025agentrl} specifically target the stability of multi-turn RL, introducing mechanisms like cross-policy sampling and step-wise evaluation to ensure robust collaboration and credit assignment in decentralized agent networks.

%% file: sections/conclu.tex
\section{Conclusion}
This work presents OMAR: One Model, All Roles , a generalizable reinforcement learning framework for developing AI social intelligence through multi-turn, multi-agent self-play. We extend the paradigm of self-play from structured games like Go to group conversations in natural languages, allowing a single model to role-play all participants to achieve collective social goals. Our evaluations in SOTOPIA environments demonstrate that fine-grained social behaviors, such as empathy, persuasion, and compromise, can emerge from multi-agent, multi-turn conversations with end-of-episode rewards, and that training models on competitive scenarios like Werewolf can also incentivize collaborative behaviors.

Looking ahead, we encourage exploration into proactive agents, where each participant can learn when and when not to speak in a group conversation appropriately. We also aim to explore more complex environments with agentic tool use beyond natural language. In summary, we present an initial exploration of conversational self-play and intelligent systems that can learn from dynamic experience and evolve in social environments.

\section{Limitations}
Our work includes certain computational approximations of social interaction, such as all roles speaking simultaneously at each turn. While all utterances are concatenated and made visible to all participants in the next turn, this setup provides a streamlined alternative to the asynchronous turn-taking typically found in real conversations. Additionally, the algorithm cannot support an unlimited number of active participants, as the participant count directly determines the batch size. Lastly, reward hacking remains a significant challenge, particularly in long multi-turn conversations, and zero-sum games where there is always a winner regardless of the conversation or reasoning qualities. We view them all as opportunities for future research into more flexible and scalable social AI agents.